\documentclass{article}
\usepackage{latexsym}

\usepackage{ijcai24}
\usepackage[T1]{fontenc}
\usepackage{times}
\usepackage{soul}
\usepackage{url}
\usepackage[hidelinks]{hyperref}
\usepackage[utf8]{inputenc}
\usepackage[small]{caption}
\usepackage{float}
\usepackage{graphicx}
\usepackage{amsmath}
\usepackage{amsthm}
\usepackage{booktabs}
\usepackage[switch]{lineno}

\usepackage{algorithm}
\usepackage{algorithmic}

\usepackage{amsfonts}       % blackboard math symbols
\usepackage{nicefrac}       % compact symbols for 1/2, etc.
\usepackage{microtype}      % microtypography
\usepackage{xcolor}         % colors
\usepackage{amssymb}

\usepackage{thmtools}
\usepackage{thm-restate}

\usepackage{multirow}
\usepackage{soul}
\usepackage{placeins}
\usepackage{subfigure}
\usepackage{stackengine}
\usepackage[symbol]{footmisc}
\usepackage{enumerate}
\usepackage{mathrsfs}

\setstackEOL{\\}
\usepackage[toc,page]{appendix}

\definecolor{lightblue}{RGB}{204,186,230}
\definecolor{citeblue}{RGB}{32,67,192}

\newtheorem{lemma}{Lemma}

\newtheorem{assumption}{Assumption}

\newtheorem{proposition}{Proposition}
\newtheorem{theorem}{Theorem}
\newtheorem{definition}{Definition}

\theoremstyle{remark}

\newcommand{\E}{\mathbb{E}}

\newcommand{\R}{\mathbb{R}}
\newcommand{\cS}{\mathcal{S}} 
\newcommand{\cA}{\mathcal{A}}

\newcommand{\cM}{\mathcal{M}}

\renewcommand{\epsilon}{\varepsilon}
\renewcommand{\phi}{\varphi}

\DeclareMathOperator*{\argmax}{arg\,max}
\DeclareMathOperator*{\argmin}{arg\,min}

\title{MICRO: Model-Based Offline Reinforcement Learning with a Conservative Bellman Operator}

\author{
	Xiao-Yin Liu$^{1,2}$\and
	Xiao-Hu Zhou$^{1,2}$ \footnote{Corresponding Authors.}\and
	Guotao Li$^{1,2}$\and
	Hao Li$^{1,2}$\and
	Mei-Jiang Gui$^{1,2}$\and
	Tian-Yu Xiang$^{1,2}$\and
	De-Xing Huang$^{1,2}$\And
	Zeng-Guang Hou$^{1,2\text{ }*}$
	\affiliations
	$^1$State Key Laboratory of Multimodal Artificial Intelligence Systems Institute of Automation, Chinese Academy of Sciences, Beijing 100190, China\\
	$^2$The School of Artificial Intelligence University of Chinese Academy of Sciences, Beijing 100049, China
	\emails
	liuxiaoyin2023@ia.ac.cn,
	xiaohu.zhou@ia.ac.cn,
	zengguang.hou@ia.ac.cn
}
\begin{document}

	\maketitle
	\begin{abstract}
		Offline reinforcement learning (RL) faces a significant challenge of distribution shift. Model-free offline RL penalizes the $Q$ value for out-of-distribution (OOD) data or constrains the policy closed to the behavior policy to tackle this problem, but this inhibits the exploration of the OOD region. Model-based offline RL, which uses the trained environment model to generate more OOD data and performs conservative policy optimization within that model, has become an effective method for this problem. However, the current model-based algorithms rarely consider agent robustness when incorporating conservatism into policy. 
		Therefore, the new \textbf {\underline{m}}odel-based offl\textbf {\underline{i}}ne algorithm with a \textbf {\underline{c}}onse\textbf{\underline{r}}vative Bellman \textbf {\underline{o}}perator (MICRO) is proposed. This method trades off performance and robustness via introducing the robust Bellman operator into the algorithm. Compared with previous model-based algorithms with robust adversarial models, MICRO can significantly reduce the computation cost by only choosing the minimal $Q$ value in the state uncertainty set. Extensive experiments demonstrate that MICRO outperforms prior RL algorithms in offline RL benchmark and is considerably robust to adversarial perturbations. 
	\end{abstract}
	
	\section{Introduction}
\label{sec:introduction}
Reinforcement learning (RL) has achieved state-of-the-art performance on numerous control decision-making tasks \cite{singh2022reinforcement,yu2021reinforcement}. However, the trial-and-error procedure in which the agent interacts with the environment directly can be costly and unsafe in some scenarios. Offline RL effectively overcomes this problem by learning policy from static dataset \cite{levine2020offline}.

Offline RL suffers from the distribution shift between the learned policy and the behavior policy in the offline dataset due to limited coverage of state-action pairs. To tackle this issue, model-free offline RL algorithms either penalize value function for out-of-distribution (OOD) state-action pairs \cite{kumar2020conservative,kostrikov2021offline} or constrain the policy closed to the behavior policy directly \cite{kumar2019stabilizing,fujimoto2019off}. However, inhibition of exploration for OOD region hinders the improvement of agent performance.

Model-based offline RL (MBORL) trains the environment model using offline dataset and optimizes the policy based on the constructed environment model. Compared with model-free methods, MBORL can perform better to state not occur in offline dataset via generating more model data \cite{yu2020mopo}. However, due to the gap between the estimated and true environment, the generated model data is unreliable, which degrades agent performance. Therefore, conservatism is incorporated into the algorithm through different means.

One type is applying reward penalties for OOD data by uncertainty quantification directly \cite{yu2020mopo,kim2023model,sun2023model} or value function regularization indirectly \cite{yu2021combo,liu2023domain}. Another type employs a robust adversarial training framework to enforce conservatism by training an adversarial dynamics model \cite{rigter2022rambo,bhardwaj2023adversarial}. However, the dynamics model of the second type is constantly updated during the policy learning process, which demands more computation costs. 

Due to the gap between simulation to reality and disturbances in the real world, a robust policy is a necessity in many real-world RL applications.
However, the current MBORL algorithms rarely consider agent robustness when improving agent performance. To tackle the gap between the simulation (source) and actual (target) environment, the robust RL based on Robust Markov Decision Process (RMDP) \cite{nilim2005robust} optimizes policy under the worst possible environment \cite{pinto2017robust,kamalaruban2020robust,panaganti2022robust}. Notice that, this is consistent with the core issue of MBORL: the gap between the estimated and the actual environment. Therefore, \emph{is it possible to introduce robust RL into MBORL as a way to improve agent performance and guarantee robustness?}

\begin{figure}
	\centering
	\includegraphics[width=8.5cm]{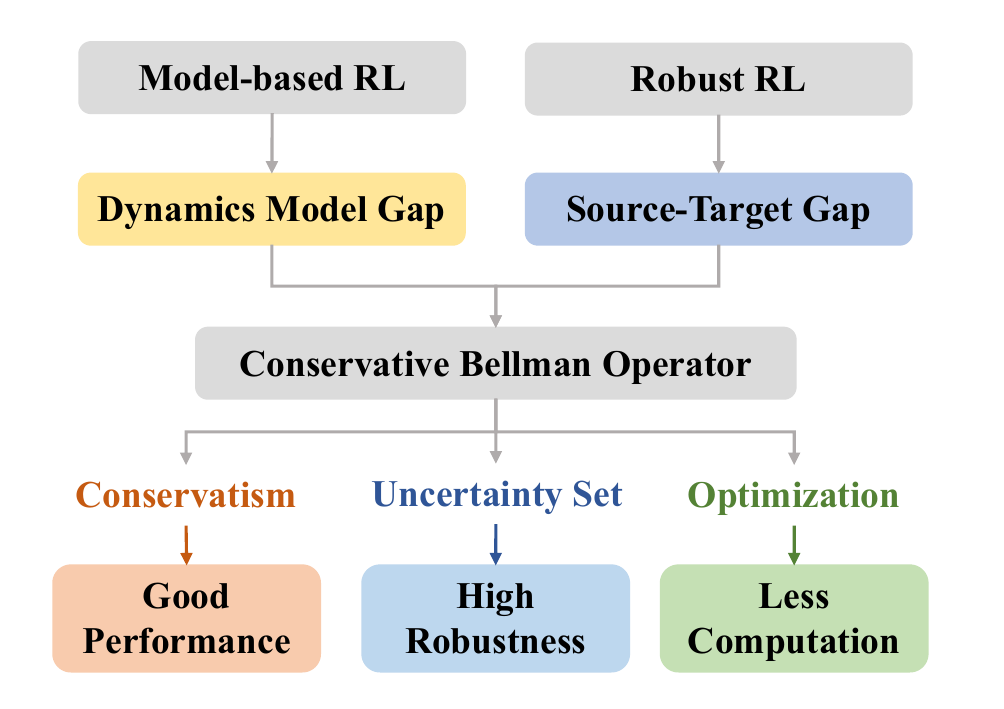}
	\caption{Conceptual illustration of model-based offline RL with a conservative Bellman operator.}
	\label{fig1}
\end{figure}

Motivated by the above question, a new MBORL algorithm is proposed, named MICRO, where a new conservative Bellman operator is designed to incorporate conservatism and guarantee robustness. Compared with \cite{rigter2022rambo,bhardwaj2023adversarial}, MICRO only needs to choose the minimal $Q$ value in the state uncertainty set, which significantly reduces computation cost. Fig. \ref{fig1} illustrates the framework of MICRO conceptually. The main contributions are summarized as follows: 

\begin{itemize}
	\item MICRO, a novel and theoretically grounded method, is the first MBORL algorithm that trades off performance and robustness by introducing robust Bellman operator.
	\item A conservative Bellman operator, which combines standard and robust Bellman operators, is designed to impose conservatism and guarantee agent robustness.
	\item Compared with previous MBORL algorithms, MICRO can guarantee agent robustness with less computation cost while improving performance.
\end{itemize}

In this paper, the theoretical and experimental results show that MICRO has the guarantee of a robust policy improvement and outperforms current state-of-the-art algorithms on the D4RL dataset benchmark. Furthermore, MICRO achieves better robustness on different adversarial attacks.
	
\section{Related Work}\label{sec:Related work}
\textbf{Model-free offline RL:} Online RL algorithms perform poorly in offline RL due to value overestimation and distribution shift. Existing model-free algorithms typically adopt two methods: policy constraint and value regularization. The goal of policy constraint is to restrict the learned policy close to the behavior policy \cite{kumar2019stabilizing,fujimoto2019off}. In contrast, value regularization methods underestimate the value for OOD data by conservatively estimating $Q$ values \cite{kumar2020conservative,cheng2022adversarially} or assigning penalization based on the $Q$ function uncertainty \cite{an2021uncertainty,bai2021pessimistic}. However, the root of challenge for offline RL stems from limited data coverage. Model-free offline RL algorithms learn policy only from offline dataset, which hinders agent exploration.

\noindent \textbf{Model-based offline RL:} Model-based offline RL algorithms train a dynamics model from the offline dataset and utilize the model to broaden data coverage. However, there is a gap between the learned and actual dynamics model due to the inadequate offline dataset. The conservatism should be incorporated into algorithms to prevent agent from areas where the prediction of the learned dynamic model is unreliable. One method is to penalize the reward by uncertainty quantification \cite{yu2020mopo,kim2023model,sun2023model}. Another method is to enforce small $Q$-value for imprecise model data generated by the dynamics model \cite{yu2021combo,li2023crop,liu2023domain}. Different from the above methods, RAMBO \cite{rigter2022rambo} and ARMOR \cite{bhardwaj2023adversarial} incorporated conservatism by  training the robust adversarial model. However, the above two methods demand high computation cost and perform poorly compared with current SOTA methods.

\noindent \textbf{Robust RL:} The research line of robust RL mainly includes two parts. One line of works focus on the adversarial learning. They applied adversarial attacks to states or actions during policy learning \cite{pinto2017robust,pattanaik2018robust,tessler2019action}. Another line of works are based on RMDP, which optimize policy under the worst-case environment. The related studies are centered around how to solve minimization problem under uncertainty set \cite{wang2021online,badrinath2021robust,panaganti2022robust}. Similar to \cite{kuang2022learning}, the robust RL in MICRO ultimately only needs to consider the state uncertainty set, which reduces computation complexity.

\noindent \textbf{This work:} The works that are closest to our work MICRO are by RAMBO \cite{rigter2022rambo} and ARMOR \cite{bhardwaj2023adversarial} that introduces the robust adversarial idea into algorithm. MICRO is significantly different from these in the following ways: 1) MICRO incorporates conservatism into policy by introducing robust Bellman operator, instead of training the adversarial model like RAMBO and ARMOR. 2) MICRO directly selects minimal $Q$ value in state uncertainty set rather than constantly updating environment model during policy learning, which consumes less computation time. 3) MICRO learns the robust policy against environment perturbations and adversarial attacks, which RAMBO and ARMOR fail to consider. \emph{To the best of our knowledge, MICRO is the first work that incorporates the robust Bellman operator into model-based offline RL, and guarantees the performance and robustness of the learned policy with less computation.}
	\section{Preliminaries}\label{sec:Preliminaries}

\textbf{Markov Decision Process and Offline RL:} An Markov Decision Process (MDP) is defined by the tuple,  $\mathcal{M}=(\mathcal{S}, \mathcal{A}, r, T_{\mathcal{M}}, \mu_{0}, \gamma)$, where $\cS$ is the state space, $\cA$ is the action space, $r: \cS\times \cA\rightarrow \R$ is the reward function in $\mathcal{M}$, $T_{\mathcal{M}}: \cS\times \cA\rightarrow \Delta({\cS})$ is the transition dynamics under $\mathcal{M}$, $\Delta({\cS})$ denotes the set of probability distribution over $\cS$, $\mu_{0}$ is the initial state distribution, and $\gamma \in(0,1]$ is the discount factor. A policy $\pi : \cS \rightarrow \Delta({\cA})$ is a conditional distribution over action given a state. The state-action value function $Q_{\mathcal{M}}^{\pi}(s,a)$ (i.e., $Q$-function) represents the expected discounted return from taking action $a$ at state $s$ and then executing $\pi$ thereafter in $\mathcal{M}$: $Q_{\mathcal{M}}^{\pi}(s,a): =\mathbb{E}_{\pi, T_\mathcal{M}}[\sum^{\infty}_{t=0} \gamma^{t} r(s_t,a_t)~|~s_{0}=s, a_{0}=a]$. The value function is defined as $V_{\mathcal{M}}^{\pi}(s):=\sum_{a \in \mathcal{A}}\pi(a~|~s)Q_{\mathcal{M}}^{\pi}(s,a)$.

The goal of RL is to optimize the policy $\pi$ that maximizes the expected discounted return $J(\mathcal{M},\pi):=\sum_{s \in \mathcal{S}}\mu_{0}(s)V_{\mathcal{M}}^{\pi}(s)$. Such policy $\pi$ can be derived from $Q$-learning, which learns the $Q$-function that satisfies the following Bellman optimal operator:
\begin{equation}
	%\hspace{-3mm}
	\label{eq_1}	
	\mathcal{T}_{\cM} Q=r(s, a)+\gamma \mathbb{E}_{s^{\prime} \sim T_{\mathcal{M}}\left(s^{\prime} \mid s, a\right)}\left[\max_{a'\in \cA} Q(s', a')\right].
\end{equation}
where the operator $\mathcal{T}_{\cM}$ is a contraction on $\mathbb{R}^{|\cS \times \cA|}$ and $Q_{\mathcal{M}}^{\star}$ is the fixed point of the above operator. Therefore, the $Q$-iteration can be defined as $Q_{k+1}=\mathcal{T}_{\cM}Q_k$, and it satisfies that $Q_k\rightarrow Q_{\mathcal{M}}^{\star}$. The $Q$-iteration uses a function class $\mathcal{Q}$, defined as $\mathcal{Q}=\{Q:\cS \times \cA \rightarrow [0,1/(1-\gamma)]\}$, to approximate $Q_{\mathcal{M}}^{\star}$. The sample-based counterpart of Bellman operator $\mathcal{T}_{\cM}$ can be expressed as $\widehat{\mathcal{T}}_{\cM} Q(s, a):=r(s, a)+\gamma [\max_{a'\in \cA}Q(s', a')]$. The $Q$-function is updated by $Q_{k+1}=\argmin_{Q\in \mathcal{Q}} \sum^{N}_{i=1}[\widehat{\mathcal{T}}_{\cM} Q_k(s, a)-Q(s,a)]^2$. The dataset $\{\left(s_{i}, a_{i}, r_{i}, s_{i}^{\prime}\right)\}_{i=1}^{N}$ can be a fixed offline dataset or replay buffer generated by the current policy interacting with the environment.

The goal of offline RL is to learn a policy which maximizes $J(\mathcal{M},\pi)$ from the fixed dataset $\mathcal{D}_{\text{offline}}=\{\left(s_{i}, a_{i}, r_{i}, s_{i}^{\prime}\right)\}_{i=1}^{N}$ collected by a behavior policy $\mu$. The discounted state  visitation distribution obtained by rolling out policy $\pi$ on $\mathcal{M}$, denoted as $d_{\mathcal{M}}^{\pi}(s):=(1-\gamma) \E_{\pi,T_{\cM}}[\sum^{\infty}_{t=0} \gamma^{t} \mathcal{P}(s_t=s|\pi,\cM)]$, where $\mathcal{P}(s_t=s|\pi,\cM)$ is the probability of reaching state $s$ at time $t$. The state-action visitation distribution is $d_{\mathcal{M}}^{\pi}(s,a)=d_{\mathcal{M}}^{\pi}(s)\pi(a|s)$. We denote the ground truth MDP as $\cM^{\star}$, so the state-action visitation distribution of a fixed dataset can be written as $d_{\mathcal{M}^{\star}}^{\mu}(s, a)$. However, since some $(s,a)$ barely occurs in $\mathcal{D}_{\text{offline}}$, implying $d_{\mathcal{M}^{\star}}^{\mu}(s, a)$ approaches zero, offline RL tends to suffer from extrapolation error.

\noindent\textbf{Robust MDP and Robust RL:} The Robust Markov Decision Process (RMDP) considers the discrepancy between simulation and real environments, and gives an \textit{uncertainty set} $\mathcal{M}_{\epsilon}$ of possible environments. Following \cite{abdullah2019wasserstein,kuang2022learning}, \textit{Wasserstein Distance} of order $p$ is used to describe uncertainty set $\mathcal{M}_{\epsilon}$, which is defined as 
\begin{equation}
	\label{eq_2}	
	\mathcal{M}_{\epsilon} = \left\{\widehat{\mathcal{M}}: \mathbb{W}_{p}\left(T_{\mathcal{M}^{\star}}(\cdot \mid s, a), T_{\widehat{\mathcal{M}}}(\cdot \mid s, a)\right)<\epsilon \right\}.
\end{equation}
where $T_{\widehat{\mathcal{M}}}(\cdot\mid s, a)$ and $T_{\mathcal{M}^{\star}}(\cdot\mid s, a)$ are the transition dynamics under simulation and real environment, respectively. If the reward function is unknown, since the analysis is similar to the above, we can concatenate the reward onto the state. $\mathbb{W}_{p}$ is $p^{th}$-order Wasserstein distance given by
\begin{equation}
\label{W}	 
\hspace{-2mm}
\mathbb{W}_{p}(T_{\widehat{\mathcal{M}}},T_{\mathcal{M}^{\star}})= \inf _{\kappa \in \mathbf{K}(T_{\widehat{\mathcal{M}}},T_{\mathcal{M}^{\star}})}\bigg\{\int_{\cS \times \cS} d(x, y)^{p} d\kappa \bigg\}^{\frac{1}{p}}.
\end{equation}
where $\mathbf{K}(T_{\widehat{\mathcal{M}}},T_{\mathcal{M}^{\star}})$ denotes the set of couplings between $T_{\widehat{\mathcal{M}}}$ and $T_{\mathcal{M}^{\star}}$, $d(\cdot,\cdot): \cS\times \cS\rightarrow \R_+$ is a distance metric for $\cS$ \cite{hou2020robust}. 

The objective of robust RL is to find the optimal robust policy that maximizes the expected discounted return against the worst possible environment in uncertainty set $\mathcal{M}_{\epsilon}$. The \textit{robust value function} $V^{\pi}$ and the \textit{robust $Q$-function} $Q^{\pi}$ are defined as $V^{\pi}(s) = \inf_{\mathcal{M} \in \mathcal{M}_{\epsilon}} V_{\mathcal{M}}^{\pi}(s), Q^{\pi}(s,a) = \inf_{\mathcal{M} \in \mathcal{M}_{\epsilon}} Q_{\mathcal{M}}^{\pi}(s,a)$, respectively \cite{nilim2005robust}. By incorporating uncertainty set into Bellman operator, the \textit{robust Bellman operator} can be denoted  as \cite{iyengar2005robust}
\begin{equation}
	\label{eq_3}
	\begin{aligned}
	\mathcal{T}_{\cM_{\epsilon}}Q = r+ \gamma \inf_{\cM \in \cM_{\epsilon}} \mathbb{E}_{s^{\prime} \sim T_{\mathcal{M}}} \left[ \max_{a'\in \cA} Q(s', a')\right]. 
	\end{aligned}
\end{equation}
Since the above formulation involves a minimization over the uncertainty set $\cM_{\epsilon}$, RMDP is much more computationally intensive compared to standard MDP, which poses a great computational challenge.

\noindent \textbf{Model-Based Offline RL:} MBORL uses the offline dataset $\mathcal{D}_{\text {offline }}$ to estimate transition dynamics model $T_{\widehat{\cM}}(s^{\prime}~|~s, a)$, which is typically trained via maximum likelihood estimation (MLE) : $\min_{T_{\widehat{\cM}}} \mathcal{L}_{\widehat{\cM}}:=\mathbb{E}_{(s, a,s^{\prime})\sim \mathcal{D}_{\text {offline }}}[-\log T_{\widehat{\cM}}(s^{\prime}~|~ s, a)]$. In the following part, we assume that the reward function $r$ is known, although the $r$ can be learned together with dynamics model if unknown. Then, the estimated MDP $\widehat{\mathcal{M}}=(\mathcal{S}, \mathcal{A}, r, T_{\widehat{\mathcal{M}}}, \mu_{0}, \gamma)$ can be constructed to approximates the true MDP $\cM^{\star}$. The MBORL methods use the estimated MDP $\widehat{\mathcal{M}}$ to generate model data $\mathcal{D}_{\text {model }}$ through $h$-step rollouts, which broadens the data coverage and mitigates the problem of extrapolation errors. The current MBORL algorithms use an augmented dataset $\mathcal{D}_{\text {offline }} \cup \mathcal{D}_{\text {model }}$ to train the policy, where each data is sampled from the model data $\mathcal{D}_{\text {model }}$ with the probability $f,f\in [0,1]$, and from the offline data $\mathcal{D}_{\text {offline }}$ with probability $1-f$. However, due to the gap between the estimated and true MDP, some generated model data are inaccurate, leading to poor performance of agent. Therefore, how to strike a balance between accurate offline data and inaccurate model data becomes the key to improving the performance of MBORL algorithms.

	\section{Problem Formulation}\label{sec:Problem Formulation}
To address the MBORL problem proposed in Section \ref{sec:Preliminaries}, conservative policy optimization is performed within that model \cite{kim2023model,yu2021combo,yu2020mopo}. Different from the above methods, ~\cite{rigter2022rambo} and ~\cite{bhardwaj2023adversarial} introduced the robust adversarial idea into MBORL. They both incorporate the policy learning information into the dynamics model learning to train the robust dynamics model, which effectively improves agent performance. 

However, there are the following drawbacks for the above two methods: \hypertarget{dra1}{1)} The dynamics model is constantly updated during the policy learning process, which needs much more computation cost. \hypertarget{dra2}{2)} They only consider the agent policy improvement and ignore the agent robustness. Therefore, we want to design an MBORL algorithm with good performance, strong robustness and low computation cost. Before this, we first make Assumption \ref{ass_1} that is widely used in the offline RL literature \cite{chen2019information,xie2021bellman,wang2021statistical}. This assumption states that there must exist a function $Q \in \mathcal{Q}$ which can well-approximate $Q_{\cM}^{\star}$ under the operator $\mathcal{T}_\cM$.
\begin{assumption}[Approximate realizability]
	\label{ass_1}
	Let $d^{\pi}\in \Delta(\cS \times \cA)$ be the arbitrary data distribution. Then, for any policy $\pi \in \Pi$, $\inf_{Q \in \mathcal{Q}}\mathbb{E}_{(s,a)\sim d_{\cM}^{\pi}}[(Q-\mathcal{T}_{\cM}Q)^2]<\epsilon_{r}$ holds.
\end{assumption}
Inspired by the goal of robust RL to train a robust policy that can be against the gap between the simulation and the actual environment, we introduce robust RL to MBORL as a way to cope with the gap between the trained and the actual environment in MBORL. \cite{rigter2022rambo} gave the following proposition, proven by \cite{uehara2021pessimistic}, which shows that the performance of any policy in the true MDP is at least as good as in the worst-case MDP with high probability.
\begin{proposition}[Pessimistic value function]
	\label{pro_1}
	Let $\cM^{\star}$ be the true MDP and $\cM_{\epsilon}$ be the uncertainty set of MDP. Then with high probability, for any policy $\pi$, $\inf_{\cM \in \cM_{\epsilon}}V_{\cM}^{\pi}\leq V_{\cM^{\star}}^{\pi}$ holds.
\end{proposition}
Proposition \ref{pro_1} states that the value under the worst MDP is a lower bound on that in the real MDP, which guarantees the conservatism of value function in uncertainty set $\cM_{\epsilon}$. Therefore, we combine robust RL and MBORL directly and propose a new conservative Bellman operator $\mathcal{T}Q(s, a)$.
\begin{definition}
	The conservative Bellman operator $\mathcal{T}$ is defined as
	\begin{equation*}
		\label{eq_4}
		\mathcal{T}Q = \left\{\begin{aligned}
			&r+\gamma \mathbb{E}_{s^{\prime} \sim T_{\mathcal{M}}}\left[\max_{a'\in \cA} Q(s', a')\right], \quad \qquad\cM=\cM^{\star}.\\
			&r + \gamma \inf_{\cM \in \cM_{\epsilon}} \mathbb{E}_{s^{\prime} \sim T_{\mathcal{M}}} \left[ \max_{a'\in \cA} Q(s', a')\right], \cM=\widehat{\cM}.
		\end{aligned}\right.
	\end{equation*}
\end{definition}
The above operator expresses: 1) When the data are accurate offline data collected from the true MDP $\cM^{\star}$, the standard Bellman operator is applied for the computation. 2) When the data are inaccurate model data collected from the estimated MDP $\widehat{\cM}$, the robust Bellman operator is used to perform conservative computation. The conservative Bellman operator incorporates conservatism into the algorithm, which reduces training risk from model errors. Meanwhile, this operator can guarantee robustness by optimizing policy under worst-case MDP. Compared with \cite{rigter2022rambo} and \cite{bhardwaj2023adversarial}, this method only needs to consider the worst MDP and does not require updating the dynamics model constantly, which significantly reduces computation cost. Next, we theoretically analyze dynamic programming properties of $\mathcal{T}$ in the tabular MDP setting and give the following proposition.
\begin{proposition}[$\gamma$-contraction mapping operator]
	\label{pro_2}
	Let $ \|\cdot\|_{\infty}$ be the $\mathcal{L}_{\infty}$ norm and $(\mathbb{R}^{|\cS \times \cA|},\|\cdot\|_{\infty})$ is complete space. Then, the conservative Bellman operator $\mathcal{T}$ is a $\gamma$-contraction mapping operator, i.e. $\|\mathcal{T} Q_{1}-\mathcal{T} Q_{2}\|_{\infty}\leq \gamma\|Q_{1}-Q_{2}\|_{\infty}$.
\end{proposition}
The proof of Proposition \ref{pro_2} is deferred to Appendix \ref{proof_pro2}. 
According to the fixed point theorem, any initial $Q$ function can converge to a unique fixed point by repeatedly applying the conservative Bellman operator $\mathcal{T}$. Note that the ground truth MDP $\cM^{\star}$ is unknown in Eq. \eqref{eq_2}. Here, considering $\cM^{\star} \in \cM_{\epsilon}$ holds with high probability, the MDP $\widehat{\cM}_{\text{MLE}}$ trained by MLE given offline dataset $\mathcal{D}_{\text{offline}}$ is employed to approximate the true MDP $\cM^{\star}$. 

However, the computation of operator $\mathcal{T}$ also faces the following crucial challenge.
\label{cha_2} It is infeasible to consider all MDPs in the uncertainty set $\cM_{\epsilon}$ for minimization in Eq. \eqref{eq_3}. \textit{How to solve the optimization over uncertainty set $\cM_{\epsilon}$?} For this challenge, \cite{panaganti2022robust} gave the dual reformulation of RMDP and solved the dual problem using empirical risk minimization. However, they introduced a new network that is updated constantly in the solving process, which also demands much computation cost. Therefore, the following section gives a solution for this challenge and derives how to further incorporate robust RL into MBORL.
	\section{Methodology}
\label{method}
\subsection{Reformulation for Conservative Bellman Operator}
Following the previous works \cite{kuang2022learning,blanchet2019quantifying}, the dual reformulation of optimization, $\inf_{\cM \in \cM_{\epsilon}} \mathbb{E}_{s^{\prime} \sim T_{\mathcal{M}}\left(s^{\prime} \mid s, a\right)} [ \max_{a'\in \cA} Q(s', a')]$, is given below.
\begin{proposition}
	\label{pro_3}
	Let $\cM_{\epsilon}$ be the uncertainty set defined by \eqref{eq_2}. Then, for any $Q:\cS \times \cA \rightarrow [0,1/(1-\gamma)]$, the term $\inf_{\cM \in \cM_{\epsilon}} \mathbb{E}_{s^{\prime} \sim T_{\mathcal{M}}\left(s^{\prime} \mid s, a\right)} [ \max_{a'\in \cA} Q(s', a')]$ can be equivalently written as
	\begin{equation*}
	\label{eq_5}
		\sup_{\lambda \geq 0}\left \{ \mathbb{E}_{s'\sim T_{\cM^\star}} \left[
		\inf_{\bar{s} \in \cS} \left(\max_{a'\in \cA}Q(\bar{s},a')+\lambda d(s',\bar{s})^p  \right)
		-\lambda\epsilon\right]\right\}.
	\end{equation*}
\end{proposition}
The proof can be found in Appendix \ref{proof_pro3}. Note that, the expection only relies on the true MDP $\cM^\star$. This avoids the demand to consider all MDP in $\cM_{\epsilon}$. However, the above equation involves minimum and maximum optimization, which is also difficult to solve. The above equation can be reduced to the following optimization problem: $\max_{\lambda}\inf_{x \in \cS}\{q(x)+\lambda(d(x)-\epsilon)\},\lambda\geq 0$. The above optimization is essentially a Lagrange dual problem for the following optimization: $\min_{x}q(x), d(x)\leq \epsilon$. Therefore, the above equation is equivalent to the following form
\begin{equation}
	\label{eq_6}
	\min_{\bar{s}} \mathbb{E}_{s'\sim T_{\cM^\star}}\left[\max_{a'\in \cA}Q(\bar{s},a')\right] \quad s.t. \quad d(\bar{s},s')^p\leq \epsilon.
\end{equation}
Let $U_{\epsilon}(s')$ be the uncertainty set of $s'$, denoted as $U_{\epsilon}(s')=\{\bar{s}\in \cS:d(\bar{s},s')^p\leq \epsilon\}$. Then, the Eq. \eqref{eq_6} can be written as $\mathbb{E}_{s'\sim T_{\cM^\star}}\{\inf_{\bar{s} \in U_{\epsilon}(s')}[\max_{a'\in \cA}Q(\bar{s},a')]\}$. Since the $\widehat{\cM}_{\text{MLE}}$ is used to approximate the true $\cM^{\star}$, the conservative Bellman operator $\mathcal{T}$ can be expressed as 
\begin{equation*}
	\label{eq_7}
	\mathcal{T}Q = \left\{\begin{aligned}
		&r+\gamma \mathbb{E}_{s^{\prime} \sim T_{\mathcal{M}}}\left[\max_{a'\in \cA} Q(s', a')\right], \quad \qquad \cM=\cM^{\star}.\\
		&r + \gamma \mathbb{E}_{s'\sim T_{\cM}}\left\{\inf_{\bar{s} \in U_{\epsilon}(s')}\left[\max_{a'\in \cA}Q(\bar{s},a')\right]\right\},\widehat{\cM}_{\text{MLE}}.
	\end{aligned}\right.
\end{equation*}
The operator $\mathcal{T}$ performs conservative estimation for model data by choosing the worst-case state $\bar{s}$ from $U_{\epsilon}(s')$, which also improves the agent robustness. Although the above operator still faces the optimization dilemma, it is easier to analyze the state compared with MDP. Next, the best function mapping $f(s,a)$ is chosen from the function class $\mathcal{F}=\{f:\cS \times \cA \rightarrow \mathbb{R}\}$ that satisfies the following form 
\begin{equation}
	\label{eq_8}
	\begin{aligned}
	\mathcal{T}Q=r+\gamma \mathbb{E}_{s^{\prime} \sim T_{\mathcal{M}}}\left[\max_{a'\in \cA} Q(s', a')-\beta f(s,a)\right].
	\end{aligned}
\end{equation}
where $f(s,a)$ is the penalty term, $\beta$ is a tuning coefficient. Since $f(s,a)=0$ when $\cM=\cM^\star$, we only need to consider the solution for $f\in \mathcal{F}$ under the estimated MDP $\widehat{\cM}_{\text{MLE}}$. Theoretically, the empirical risk minimization approach is applied to find $f$, denoted as
\begin{equation}
	\label{eq_9}
	\begin{aligned}
	f=&\argmin_{f\in \mathcal{F}} \mathbb{E}_{(s,a,s')\sim \mathcal{D}_{\text{model}}}\bigg \{\bigg(\max_{a'\in \cA} Q(s', a')\\
	&-\inf_{\bar{s} \in U_{\epsilon}(s')}\left[\max_{a'\in \cA}Q(\bar{s},a')\right]\bigg)-\beta f(s,a)\bigg \}^2.
	\end{aligned}
\end{equation}
where $\mathcal{D}_{\text{model}}$ is the dataset collected from $\widehat{\cM}_{\text{MLE}}$. Due to uncertainty in the model parameters, the ensemble of $N$ dynamics models,  $\{T^{i}_{\phi}=\mathcal{N}(\mu^{i}_{\phi},\sigma^{i}_{\phi})\}_{i=1}^N$, is trained similar to \cite{yu2020mopo,janner2019trust}, each of which is a neural network that outputs a Gaussian distribution. Notice that there is a bias in the states predicted by each dynamics model, so the set $\mathcal{X}(s,a) = \{s'|s'\sim T^{i}_{\phi}(s'|s,a),i=1,2,...,N\}$ is applied to approximate $U_{\epsilon}(s')$. Further, the following equation is adopted to approximate $f(s,a)$:
\begin{equation}
	\label{eq_10}
	f(s,a)=\max_{a'\in \cA} Q(s', a')-\inf_{\bar{s}\in \mathcal{X}(s,a)}\left[\max_{a'\in \cA} Q(\bar{s}, a')\right]\text{,}
\end{equation}
where $s'\sim \sum_{i=1}^{N}T^{i}_{\phi}(s'|s,a)/N$. Since $\mathcal{X}(s,a)$ still fails to cover all region of the state uncertain set, the coefficient $\beta$ is introduced in Eq. \eqref{eq_8} to adjust the value. From this, the $f(s,a)$ also reflects the gap of $Q$-function value between the worst-case and average level. Obviously, Eq. \eqref{eq_8} also satisfies $\gamma$-contraction mapping, which guarantees the iteration convergence.

Note that $f(s,a)$ can be regarded as a type of uncertainty quantification like MOPO \cite{yu2020mopo}. However, $f(s,a)$ contains more robustness information. This term trade off performance and robustness. For example, if $\epsilon_1 \leq \epsilon_2$, then $U_{\epsilon_1}(s')\subseteq U_{\epsilon_1}(s')$ holds, thereby $\inf_{\bar{s} \in U_{\epsilon_1}(s')}[F(\bar{s})]\geq \inf_{\bar{s} \in U_{\epsilon_2}(s')}[F(\bar{s})]$ holds. Therefore, as $\epsilon$ increases, the penalty $f(s,a)$ for model data increases, which means robustness increases while performance is difficult to improve. The $\epsilon$ for different data $(s,a)$ is adjusted dynamically by model uncertainty indirectly in Eq. \eqref{eq_10}, instead of the fixed value. This is beneficial to trading off the performance and robustness.

\subsection{Theoretical Analysis}
The theoretical guarantees on \emph{Robust Policy Improvement} property of MICRO are given in this section. First, the policy concentrability coefficient is introduced below \cite{bhardwaj2023adversarial,uehara2021pessimistic}.
\begin{definition}[Policy concentrability coefficient]
	We define $\mathscr{C}_{\cM_{\epsilon}}(\pi)$ to measure how well model errors transfer between the visitation distribution $d_{\widehat{\cM}}^{\pi}$ under estimated MDP $\widehat{\cM}$ and policy $\pi$ and offline data distribution $d_{\cM^\star}^{\mu}$,
	\begin{equation*}
		\label{def}
		\mathscr{C}_{\cM_{\epsilon}}(\pi) = \sup_{\widehat{\cM} \in \cM_{\epsilon}}\frac{\mathbb{E}_{(s,a) \sim d_{\widehat{\cM}}^{\pi}}\left[\text{\rm TV}\Big(T_{\mathcal{M}^{\star}}, T_{\widehat{\mathcal{M}}}\Big)\right]}{\mathbb{E}_{(s,a) \sim d_{\cM^\star}^{\mu}}\left[\text{\rm TV}\Big(T_{\mathcal{M}^{\star}}, T_{\widehat{\mathcal{M}}}\Big)\right]}.
	\end{equation*}
\end{definition}
\noindent where TV is total variation distance, and the $\mathscr{C}_{\cM_{\epsilon}}(\pi)$ is upper bounded by $\|d_{\widehat{\cM}}^{\pi}/d_{\cM^\star}^{\mu}\|_{\infty}$\cite{uehara2021pessimistic}. It represents the discrepancy between the offline data distribution compared to the visitation distribution under policy $\pi$.

The optimal policy $\pi^\star$ of standard RL can be obtained by maximizing $J(\mathcal{M}, \pi)= \mathbb{E}_{(s, a) \sim d_{\mathcal{M}}^\pi(s, a)}[r(s, a)]/({1-\gamma})$. Note that in Eq. \eqref{eq_8}, the conservative Bellman opeator $\mathcal{T}$ is equivalent to penalize the reward $r(s,a)$, i.e. $\widehat{r}(s,a)=r(s,a)-\gamma\beta f(s,a)$. Therefore, the output policy ${\pi}^\dag$ of MICRO is expressed as
\begin{equation}
	\label{eq_12}
	{\pi}^\dag =\argmax_{\pi \in \Pi}\bigg\{J(\cM_f,\pi)-\frac{\gamma\beta}{1-\gamma}\mathbb{E}_{ d_{\mathcal{M}_f}^\pi}\big[f(s, a)\big]\bigg\}.
\end{equation}
where the $\cM_f=f\widehat{\cM}+(1-f)\overline{\cM}$ denotes $f$-interpolant MDP, which combines sampling error and model error. The $\overline{\cM}$ is empirical MDP that is based on sample data. The $d_{\cM_f}^\pi=fd_{\widehat{\cM}}^\pi+(1-f)d_{\overline{\cM}}^\pi$ is the mixed distribution of offline and model dataset. Since $f(s,a)=0$ when $\cM=\cM^\star$ in Eq. \eqref{eq_12}, the final learned policy ${\pi}^\dag$ can be denoted as 
\begin{equation}
	\label{eq_13}
	{\pi}^\dag =\argmax_{\pi \in \Pi}\bigg\{J(\cM_f,\pi)-\frac{f\gamma\beta}{1-\gamma}\eta(\pi)\bigg\}.
\end{equation}
where $\eta(\pi)=\mathbb{E}_{ d_{\widehat{\mathcal{M}}}^\pi(s, a)}[f(s, a)]$ denotes the expectation of the discrepancy of $Q$ value between the worst-case and average level under estimated MDP $\widehat{\cM}$. 
The Bellman operator on empirical MDP $\overline{\cM}$ suffers from sampling error but is unbiased in expectation, whereas Bellman operator on estimated MDP $\widehat{\cM}$ induces bias but no sampling error \cite{yu2021combo,liu2023domain}. 
The performance between the true MDP $\mathcal{M}^\star$ and the estimated MDP $\widehat{\mathcal{M}}$ and the empirical MDP $\overline{\mathcal{M}}$ under policy $\pi$ can be found in Lemma \ref{lem_2} and Lemma \ref{lem_3}, respectively.

\begin{algorithm}[tb]
	\caption{Model-based offline reinforcement learning with a conservative Bellman operator (MICRO)}
	\label{alg1}
	\textbf{Input}: offline dataset $\mathcal{D}_{\text{offline}}$, critic $\{Q_{\omega_i}\}_{i=1}^K$,  policy $\pi_\theta$, environment model ${T}_\phi$, and rollout length $h$.\\
	\textbf{Output}: the final policy $\pi^\dag$ learned by the algorithm MICRO.
	\begin{algorithmic}[1] %[1] enables line numbers
		\STATE Initialize policy $\pi_\theta$, target network $ Q_{\bar{\omega}_i} \! \leftarrow \! Q_{\omega_i}$, and model dataset $\mathcal{D}_{\text{model }}=\varnothing $.
		\STATE Train an ensemble dynamics model $\{{T}_\phi^i\}_{i=1}^N$ on $\mathcal{D}_{\text{offline} }$.
		\FOR {$t=1,2,\cdots, n_{\text{iter}} $}
		\STATE Generate synthetic $h$-step rollouts by ${T}_\phi^i$. Add transition data into $\mathcal{D}_{\text{model}}$.
		\STATE Draw samples $B=(s,a,s',r)$ from $\mathcal{D}_{\text{offline}} \cup \mathcal{D}_{\text{model}}$. 
		\STATE Compute $f(s,a)$ according to Eq. \eqref{eq_10}. 
		\STATE Update policy $\pi_\theta$ and critic network $Q_{\omega_i}$ using $B$ according to Eqs. \eqref{sac} and \eqref{target}.
		\IF {$t$ \% update period = 0}
		\STATE Soft update periodically: $\bar{\omega}_{i} \leftarrow \tau \omega_{i}+(1-\tau) \bar{\omega}_{i} $.
		\ENDIF
		\ENDFOR
	\end{algorithmic}
\end{algorithm}

\begin{table*}
	\small
	\renewcommand{\arraystretch}{0.8}
	\caption{Results for the D4RL tasks during the last 5 iterations of training averaged over 4 seeds. $\pm$ captures the standard deviation over seeds. Bold indicates the highest score. Highlighted numbers indicate results within 2\% of the most performant algorithm.}
	\label{d4rl}
	\centering
	%\resizebox{\linewidth}{!}
	{
		\begin{tabular}{p{0.05cm}p{0.05cm}c|c|cccc|cccc}
			\toprule[1pt] 
			\multicolumn{2}{c}{\multirow{2}{*}{\textbf{Type}}} &\multirow{2}{*}{\textbf{Environment}} &Ours & \multicolumn{4}{c|}{Model-based Offline RL baseline} & \multicolumn{4}{c}{Model-free Offline RL baseline} \\
			&& & \textbf{MICRO} & \textbf{ARMOR} & \textbf{RAMBO}&\textbf{COUNT} & \textbf{MOBILE} & \textbf{EDAC} & \textbf{ATAC} & \textbf{PBRL} & \textbf{IQL} \\
			\midrule
			\multicolumn{2}{c}{\multirow{3}{*}{\rotatebox{90}{Medium}}} & Halfcheetah & $73.9\pm1.7$ &$54.2$ & \colorbox{lightblue}{$\mathbf{77.6}$} & \colorbox{lightblue}{$76.5$} & $74.6$& 
			$65.9$ & $54.3$ & $57.9$ & $47.4$ \\
			
			&& Hopper & \colorbox{lightblue}{$106.5\pm0.8$} & $101.4$ & $92.8$& $103.6$ & \colorbox{lightblue}{$\mathbf{106.6}$} 
			
			& $101.6$ & $102.8$ & $75.3$ & $66.3$ \\
			
			&& Walker2d & $86.6\pm1.5$  & $90.7$ & $86.9$ & $87.6$ & $87.7$&
			
			\colorbox{lightblue}{$\mathbf{92.5}$} & \colorbox{lightblue}{$91.0$} & $89.6$ & $78.3$ \\
			
			\midrule
			\multirow{3}{*}{ \rotatebox{90}{Medium} } &\multirow{3}{*}{ \rotatebox{90}{Replay} }& Halfcheetah &
			
			\colorbox{lightblue}{$71.5\pm1.9$} & $50.5$ & $68.9$ & \colorbox{lightblue}{${71.5}$}&\colorbox{lightblue}{$\mathbf{71.7}$}
			
			& $61.3$ & $49.5$ & $45.1$ & $44.2$ \\
			
			&& Hopper & \colorbox{lightblue}{$\mathbf{104.6\pm0.9}$} & $97.1$ & $96.6$  & $101.7$ &\colorbox{lightblue}{$103.9$}&
			
			$101.0$ & \colorbox{lightblue}{$102.8$} & $100.6$ & $94.7$ \\
			
			&& Walker2d & \colorbox{lightblue}{$\mathbf{94.2\pm0.9}$} & $85.6$ & $85.0$  & $87.7$ &$89.9$
			
			& $87.1$ & \colorbox{lightblue}{$94.1$} & $77.7$ & $73.9$ \\
			\midrule
			\multirow{3}{*}{ \rotatebox{90}{Medium}} &\multirow{3}{*}{ \rotatebox{90}{Expert}} & Halfcheetah &
			
			\colorbox{lightblue}{$106.8\pm1.4$} & $93.5$ & $93.7$ & $100.0$& \colorbox{lightblue}{$\mathbf{108.2}$}& 
			
			\colorbox{lightblue}{$106.3$} & $95.5$ & $92.3$ &$86.7$ \\
			
			&& Hopper &\colorbox{lightblue}{$\mathbf{113.1\pm0.4}$} & $103.4$ & $83.3$ & \colorbox{lightblue}{$111.4$} &\colorbox{lightblue}{$112.6$}
			
			& {$110.7$} & \colorbox{lightblue}{$112.6$} & $110.8$ & $91.5$ \\
			
			&&Walker2d&\colorbox{lightblue}{$\mathbf{116.3\pm0.7}$}&$112.2$ & $68.3$ & $112.3$ & \colorbox{lightblue}{$115.2$} &
			
			\colorbox{lightblue}{$114.7$} & \colorbox{lightblue}{$116.3$} & $110.1$ & $109.6$ \\
			\midrule
			\multicolumn{3}{c|}{Average Score} & $\mathbf{97.1}$ & $87.6$&$83.7$ & $94.7$ & $96.7$ & $93.5$ & $91.0$ & $84.5$ & $76.9$ \\
			\bottomrule[1pt]
		\end{tabular}
	}
\end{table*}

\begin{table}
	\small
	\renewcommand{\arraystretch}{0.9}
	\caption{Comparison results of computation time (unit: seconds).}
	\label{cost}
	\centering
	{
		\begin{tabular}{ccccc}
			\toprule[1pt]
			\multirow{2}{*}{\textbf{Method}}&\multicolumn{3}{c}{\textbf{Environment}}&\multirow{2}{*}{\textbf{Average}}\\
			\cmidrule(lr){2-4}
			&Halfcheetah&Hopper&Walker2d&\\
			\midrule
			{MICRO}&$48031.4$&$45647.6$&$49271.5$&$47650.2$\\
			{RAMBO}&$133438.0$&$118020.2$&$132339.5$&$127932.7$\\
			\toprule[1pt]
		\end{tabular}
	}
\end{table}

\begin{theorem}[Robust policy improvement]
	\label{the_1}
	Under Assumption \ref{ass_1}, for any policy $\pi \in \Pi$, the policy $\pi^{\dag}$ learned by MICRO in Eq.\eqref{eq_13}, with high probability $1-\delta$, satisfies that $J(\cM^\star,\pi)-J(\cM^\star,{\pi}^\dag)$ is upper bound by
	\begin{equation*}
		\begin{aligned}
			&\underbrace{\mathcal{O}\bigg(\frac{f\gamma\beta}{1-\gamma}\bigg)\bigg[\eta(\pi)-\eta(\pi^\dag)\bigg]}_{:=\Delta_1}+\underbrace{\mathcal{O}\bigg(\frac{1-f}{1-\gamma}\bigg)\bigg[\omega(\pi)+\omega(\pi^\dag)\bigg]}_{:=\Delta_2}\\
			&+\underbrace{\mathcal{O}\bigg(\frac{fR_{max}}{(1-\gamma)^2}\sqrt{\frac{2\log \left(|\cM|/\delta\right)}{n}}\bigg)\bigg[\mathscr{C}_{\cM_{\epsilon}}(\pi)+\mathscr{C}_{\cM_{\epsilon}}(\pi^\dag)\bigg]}_{:=\Delta_3}.
		\end{aligned}
	\end{equation*}
\end{theorem}
\noindent where $\omega(\pi)=\mathbb{E}_{d_{\overline{\mathcal{M}}}^\pi(s, a)}\left[{(C_{r, \delta}+R_{\max } C_{T, \delta})}/{\sqrt{|\mathcal{D}(s, a)|}}\right]$ reflects the sampling error under policy $\pi$. The proof of Theorem \ref{the_1} can be found in Appendix \ref{proof_the1}. Theorem \ref{the_1} proves that the policy $\pi^\dag$ learned by MICRO can compete with any arbitrary policy $\pi$ with a large enough dataset \cite{cheng2022adversarially}. The term $\Delta_1$ captures the penalty discrepancy of model uncertainty between the policy $\pi$ and $\pi^\dag$. The terms $\Delta_2$ and $\Delta_3$ reflect the sampling error and model error, respectively. When the policy $\pi$ is the behavior policy $\mu$, the generated model data are more accurate under $\widehat{\cM}$ and $\mu$, i.e. $f(s,a)$ approaching zero, which leads to $\eta(\mu)-\eta(\pi^\dag)<0$. Therefor, the upper bound of $J(\cM^\star,\mu)-J(\cM^\star,{\pi}^\dag)$ is more tight.

\subsection{Practical Implementation}

In practical implementation, the value function $Q$ and the policy $\pi$ are approximated via neural networks, denoted as $Q_{\omega}$ and $\pi_{\theta}$, respectively. The algorithm MICRO, presented in algorithm \ref{alg1}, is built on model-based policy optimization (MBPO) \cite{janner2019trust} and soft actor-critic (SAC) \cite{haarnoja2018soft}. The MLE method is used to train $N$ ensemble dynamics models $\{T^{i}_{\phi}=\mathcal{N}(\mu^{i}_{\phi},\sigma^{i}_{\phi})\}_{i=1}^N$. Before agent updating, the model data is collected through $h$-step rollouts and is added into model dataset $\mathcal{D_{\text{model}}}$. Then, higher coverage dataset, $\mathcal{D}=\mathcal{D_{\text{offline}}} \cup \mathcal{D_{\text{model}}}$, are utilized to update agent. The principle of agent updates is described below
\begin{equation}
	\label{sac}
	\hspace{-2mm}
		\begin{aligned}
			& Q_{\omega_i} \leftarrow \arg \min _{\omega_i} \mathbb{E}_{s, a, s^{\prime} \sim \mathcal{D}}\left[\left(Q_{\omega_i}(s, a)-\widehat{\mathcal{T}} Q_{\omega_i}(s, a)\right)^2\right]\\
			& \pi_\theta \leftarrow \arg \max _\theta \mathbb{E}_{s,a\sim \mathcal{D}}\left[\min_{i=1,...,K}Q_{\omega_i}(s,a)-\alpha \mathcal{H}(s,a)\right].
		\end{aligned}
\end{equation}
where $\widehat{\mathcal{T}} Q_{\omega_i}(s, a)$ is the target value, $\alpha$ is the entropy regularization coefficient, $Q_{\omega_i}$ denotes the $i$-th $Q$-function, and $K$ is the number of critic. The entropy regularization term $\mathcal{H}(s,a):=\log \pi_{\theta}(a~|~s)$ is applied to prevent policy degradation \cite{haarnoja2018soft}. The minimization operation in above equation is to prevent overestimation of agent training \cite{van2016deep}. For any pairs $(s,a,r,s') \in \mathcal{D}$, the target value can be defined as
\begin{equation}
	\label{target}
	\begin{aligned}
	\widehat{\mathcal{T}} Q_{\omega_i}(s, a) = r(s,a)+&\gamma\bigg[\min_{i=1,...,K}Q_{\bar{\omega}_i}(s',a')\\
	&-\alpha \mathcal{H}(s',a')-\beta f(s,a)\bigg].
	\end{aligned}
\end{equation}
where $Q_{\bar{\omega}_i}$ represents the $i$-th target $Q$-function, and $\widehat{\mathcal{T}}$ denotes the practical conservative Bellman operator.
	\section{Experiments}
\label{sec:experiments}
This section focuses on the following questions. 
\hypertarget{que1}{\emph{\textbf{Q1}}:} How does MICRO compare with state-of-the-art RL algorithms in standard offline RL benchmarks? 
\hypertarget{que2}{\emph{\textbf{Q2}}:} How does MICRO of computation cost compare with RAMBO and ARMOR? 
\hypertarget{que3}{\emph{\textbf{Q3}}:} Does the method perform better robustness than other RL algorithms? 
We answer the above questions using D4RL benchmark \cite{fu2020d4rl} with several control tasks and datasets. More hyperparameters and implementation details are provided in Appendix \ref{detail}. The code for MICRO is available at \href{https://github.com/xiaoyinliu0714/MICRO}{github.com/xiaoyinliu0714/MICRO}.

\subsection{Benchmark Results (Q1 and Q2)}
This section compares recent model-free offline RL algorithms, such as ATAC \cite{cheng2022adversarially}, EDAC \cite{an2021uncertainty}, IQL \cite{kostrikov2021offline}, and PBRL \cite{bai2021pessimistic}, as well as model-based offline RL algorithms, including ARMOR \cite{bhardwaj2023adversarial}, RAMBO \cite{rigter2022rambo}, MOBILE \cite{sun2023model}, and Count-MORL \cite{kim2023model} (denoted as COUNT here). These methods are evaluated on Gym domain involving three environments (HalfCheetah, Hopper, and Walker2d) with three types of datasets (medium, medium-replay, medium-expert) for each environment. 

Table \ref{d4rl} gives the results of the average normalized score with standard deviation. The results for the above methods are obtained from the original papers. The table shows that MICRO outperforms prior RL methods on 4 of total 9 tasks and achieves comparable results in the remaining settings. Overall, MICRO achieves best average performance compared to previous RL algorithms. Table \ref{cost} compares computation time between MICRO and RAMBO under three environments (ARMOR is not compared, as code is not open-source). It shows MICRO reduces computation cost by nearly three times.

\subsection{Environment Parameter Perturbations (Q3)}
\begin{figure}
	\centering
	\includegraphics[width=8.7cm]{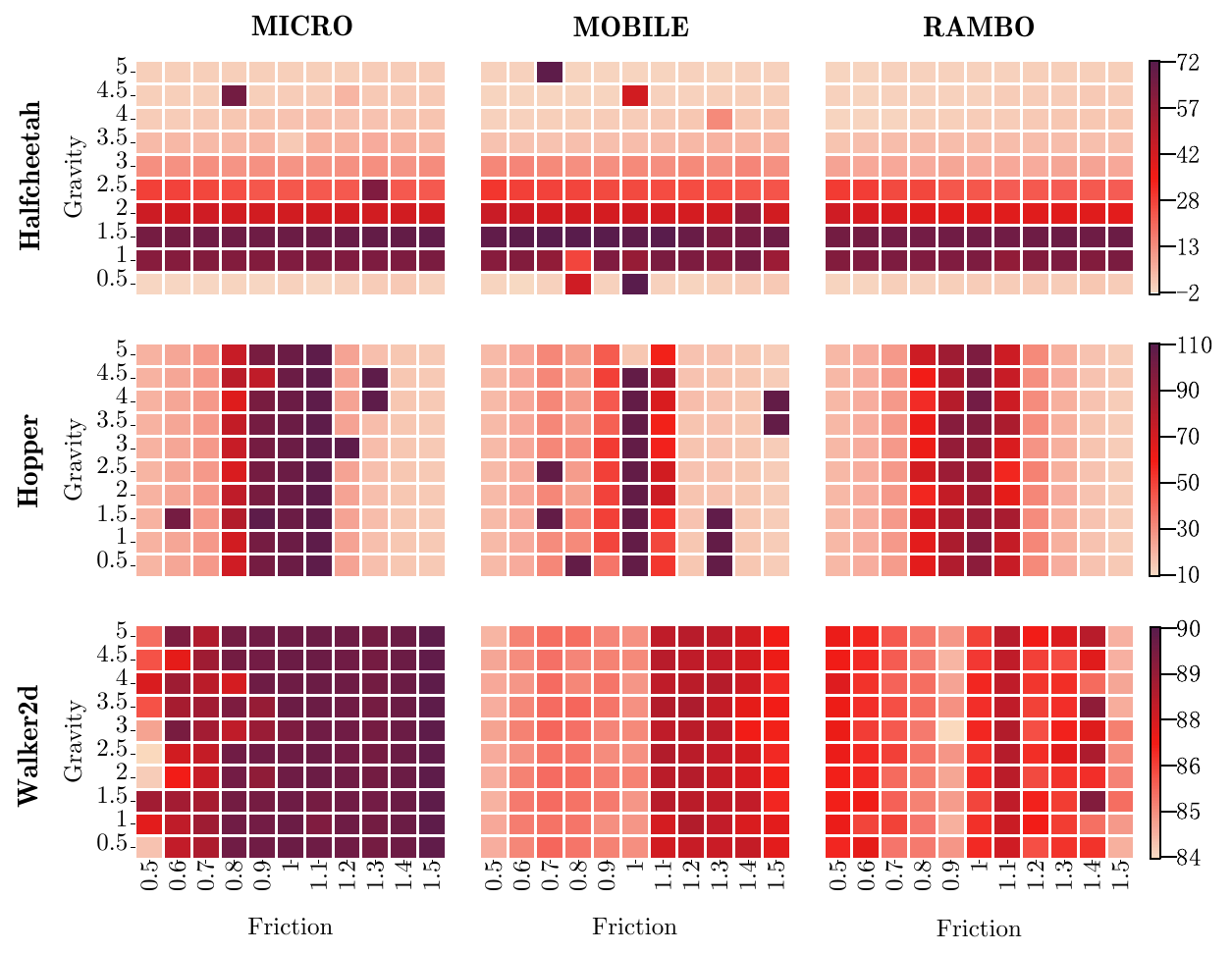}
	\caption{The performance of MICRO, RAMBO and MOBILE under environment parameters perturbation in Halfcheetah, Hopper and Walker2d environments. The gravity and friction vary from 0.5 to 5 times and 0.5 to 1.5 times the value of the simulation environment, respectively.}
	\label{fig2}
\end{figure}

In most cases, it is difficult to achieve high-fidelity simulation environment, which inevitably results in sim-to-real gaps in policy learning. The simulation environment of D4RL tasks is based on the MuJoCo physics simulator. To test the performance of method in high dynamics-gap environment, we modify the parameters of MuJoCo environment, including friction and gravity, to construct high sim-to-real gap environment following \cite{niu2022trust}. 

Fig. \ref{fig2} gives the results under various test conditions, where different gravity and friction coefficients are set for the three environments and darker colour means higher performance. The robustness of MICRO is significantly superior to MOBILE and RAMBO in sim-to-real gaps, and MICRO achieves the best generalization in Walker2d environment.

\subsection{External Adversarial Attacks (Q3)}
\begin{figure}
	\centering
	\includegraphics[width=8.7cm]{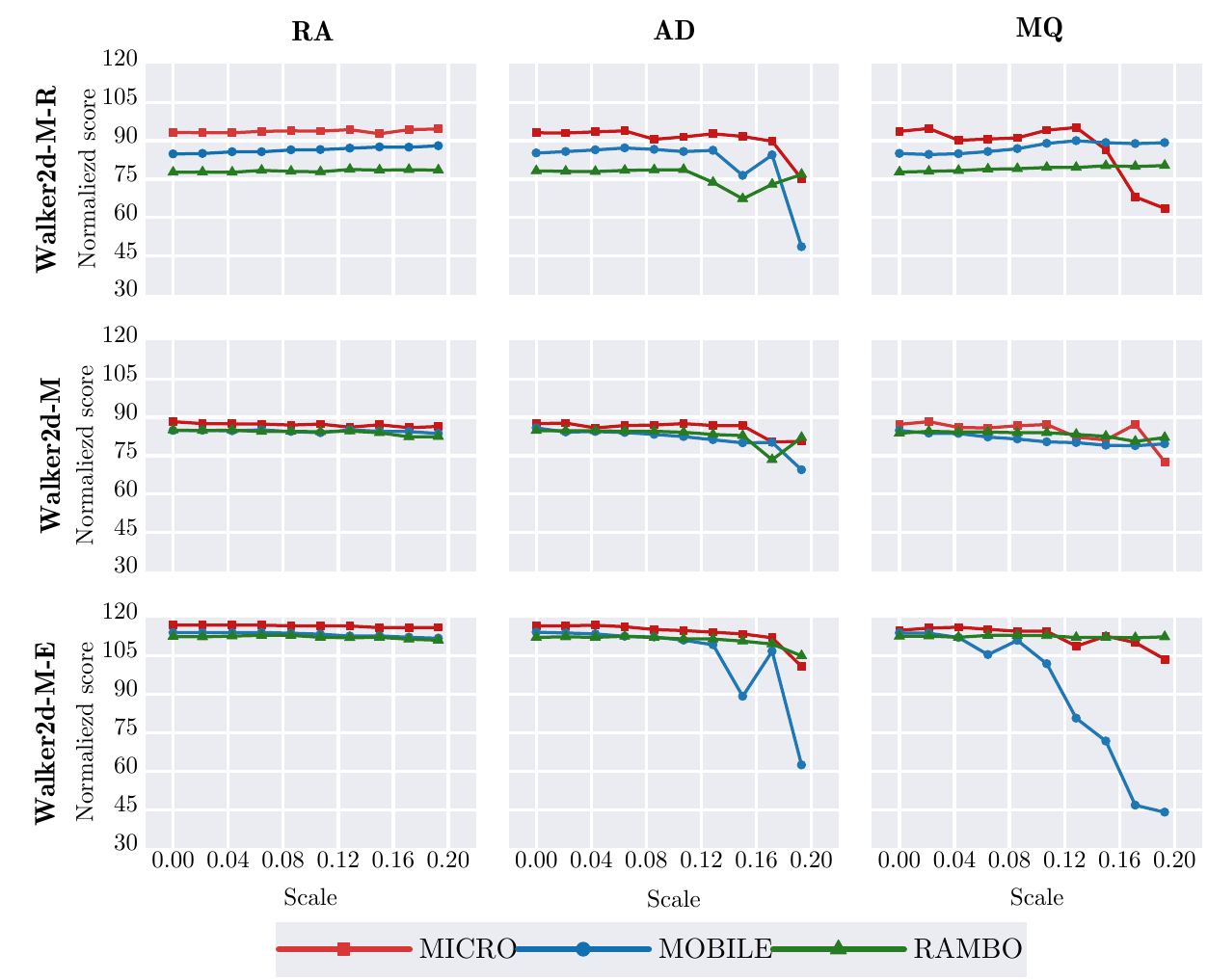}
	\caption{The performance of MICRO, MOBILE and RAMBO under attack scales range [0, 0.2] of different attack types in Walker2d environment. M, M-R and M-E are the abbreviations of Medium, Medium-Replay and Medium-Expert, respectively.}
	\label{fig3}
\end{figure}
Due to the observation errors and external disturbances within real-world scenarios, three noise perturbation methods, namely \emph{\textbf{RA}}, \emph{\textbf{AD}}, \emph{\textbf{MQ}} following \cite{yang2022rorl,zhang2020robust}, are used to test the robustness of policy. The observation states are attacked through the above perturbation in evaluation phase. Details about three perturbation types are as follows.
\begin{itemize}
	\item \emph{\textbf{RA}} randomly samples perturbed states from the ball $U_{\epsilon}(s)=\{s'\in \cS:d(s,s')^p\leq \epsilon\}$, where $\epsilon$ denotes the scale of perturbation. 
	\item \emph{\textbf{AD}} is an effective perturbation based on agent policy. It selects the state $s'$ from $U_{\epsilon}(s)$ that maximizes the gap between action distribution $\pi_\theta(\cdot |s)$ and $\pi_\theta(\cdot |s')$, i.e. $\min_{s'\in U_\epsilon(s)}-D_J(\pi_\theta(\cdot |s)||\pi_\theta(\cdot |s'))$.
	\item \emph{\textbf{MQ}} performs the perturbation based policy and $Q$-function. It finds the state $s'$ from $U_{\epsilon}(s)$ that minimizes the $Q$-function, i.e. $\min_{s'\in U_\epsilon(s)}Q(s,\pi_\theta(s'))$.
\end{itemize}
Fig. \ref{fig3} gives the result of different methods with varying perturbation scale $\epsilon$ under three perturbation types in Walker2d environment. This figure shows that MICRO achieves better robustness than other methods under \emph{RA} and \emph{AD} attacks. However, the performance of MICRO is inferior to RAMBO in large \emph{MQ} attack. This is mainly because this attack selects the minimum $Q$-value, which is more conservative for MICRO. The results in Halfcheetah and Hopper environments can be found in Appendix \ref{Robust Experiment}. Overall, MICRO greatly outperforms MOBILE in external attacks and achieves superior robustness with less time compared with RAMBO.
	\section{Conclusion}
MICRO is a promising novel model-based offline RL algorithm that introduces robust Bellman operator into method, where a conservative Bellman operator is proposed to improve performance and guarantee robustness with less computation. 
Extensive experiments demonstrate that MICRO can achieve state-of-the-art performance on standard benchmarks and better robustness under different adversarial attacks and perturbations. The main limitation of this work is the robustness slightly performs poorly under strong attacks, which should be improved in future works. MICRO is cutting-edge work in how to trade off performance and robustness in model-based offline RL field. We hope this can provide a promising direction for future research.
	
	\newpage
	\section*{Acknowledgements}
	This work was supported in part by the National Key Research and Development Program of China under 2023YFC2415100; in part by the National Natural Science Foundation of China under Grant 62222316, Grant 62373351; in part by the Youth Innovation Promotion Association of Chinese Academy of Sciences (CAS) under Grant 2020140; and in part by the CIE-Tencent Robotics X Rhino-Bird Focused Research Program.

	%\bibliography{tex/refs} 
	%\bibliographystyle{plain}
	\bibliographystyle{named}
	\bibliography{ijcai24}
	
	\onecolumn
	
	\newpage

\begin{appendices}
\section{Proof of Proposition}
\subsection{Proof of Proposition \ref{pro_2}}\label{proof_pro2}
\noindent \textit{Proof:}
Let $Q_1$ and $Q_2$ be two arbitrary $Q$-function. If the next state $s'$ is from the true MDP $\cM^{\star}$, we have
\begin{align*}
	\left\|\mathcal{T} Q_{1}-\mathcal{T} Q_{2}\right\|_{\infty}&=\gamma \max _{s, a}\left|\mathbb{E}_{s^{\prime}\sim \mathcal{T}_{\cM}}\left[\max _{a^{\prime} \sim \mathcal{A}} Q_{1}\left(s^{\prime}, a^{\prime}\right)-\max _{a^{\prime} \sim \mathcal{A}} Q_{2}\left(s^{\prime}, a^{\prime}\right)\right]\right| \\
	& \leq \gamma \max _{s, a} \mathbb{E}_{s^{\prime}\sim \mathcal{T}_{\cM}}\left|\max _{a^{\prime} \sim \mathcal{A}} Q_{1}\left(s^{\prime}, a^{\prime}\right)-\max _{a^{\prime} \sim \mathcal{A}} Q_{2}\left(s^{\prime}, a^{\prime}\right)\right| \\
	& \leq \gamma \max _{s, a}\left\|Q_{1}-Q_{2}\right\|_{\infty} =\gamma\left\|Q_{1}-Q_{2}\right\|_{\infty}.
\end{align*}
If the next state $s'$ is from the estimated MDP $\widehat{\cM}$, we have
\begin{align*}
	\left\|\mathcal{T} Q_{1}-\mathcal{T} Q_{2}\right\|_{\infty}&=\gamma \max _{s, a}\left|\inf_{\cM \in \cM_{\epsilon}}\mathbb{E}_{s^{\prime}\sim \mathcal{T}_{\cM}}\left[\max _{a^{\prime} \sim \mathcal{A}} Q_{1}\left(s^{\prime}, a^{\prime}\right)\right]-\inf_{\cM \in \cM_{\epsilon}}\mathbb{E}_{s^{\prime}\sim \mathcal{T}_{\cM}}\left[\max _{a^{\prime} \sim \mathcal{A}} Q_{2}\left(s^{\prime}, a^{\prime}\right)\right]\right| \\
	& \leq\gamma \max _{s, a}\left|\mathbb{E}_{s^{\prime}\sim \mathcal{T}_{\cM_{\eta}}}\left[\max _{a^{\prime} \sim \mathcal{A}} Q_{1}\left(s^{\prime}, a^{\prime}\right)-\max _{a^{\prime} \sim \mathcal{A}} Q_{2}\left(s^{\prime}, a^{\prime}\right)\right]\right| \quad \Leftarrow \exists \cM_{\eta} \in \cM_{\epsilon}\\
	& \leq \gamma\left\|Q_{1}-Q_{2}\right\|_{\infty}.
\end{align*}
The first inequality comes from the fact that $\min_{x_1}f(x_1)-\min_{x_2}g(x_2)\leq f(x^\star)-g(x^\star)$, where $x^\star=\argmin_x{g(x)}$. Therefore, it can be concluded that the operator $\mathcal{T}$ is a $\gamma$-contraction mapping in the $\mathcal{L}_\infty$ norm.

\subsection{Proof of Proposition \ref{pro_3}}\label{proof_pro3}
\begin{lemma}
	\label{lem_1}
	Let $D(u,v)=\inf_{\kappa \in \mathbf{K}(u,v)}\int_{(x,y)\sim \kappa}c(x,y)d\kappa$ be distance metric and $\mathcal{P}=\{P:D(P,P^\star)\leq\epsilon\}$ be uncertainty set, where $P^\star$ is baseline measure, $c(\cdot,\cdot)$ is cost function, $\mathbf{K}(u,v)$ is the set of all joint distributions with $u$ and $v$ as respective marginals. Then, the term $\inf_{P \in \mathcal{P}}\mathbb{E}_{x\sim P}[f(x)]$ is equivalent to
	\begin{equation}
		\label{aeq_1}
		\sup_{\lambda \geq 0}\left \{ \mathbb{E}_{x\sim P^\star} \left[
		\inf_{y} \left(f(y)+\lambda c(x,y)  \right)-\lambda\epsilon\right]\right\}.
	\end{equation}
\end{lemma}
\noindent The proof of lemma \ref{lem_1} can be found in \cite{blanchet2019quantifying}. Note that, when the uncertainty set is defined as Eq. \eqref{eq_2}, $c(x,y)=d(x,y)^p$ and $f(x)=\max_a Q(x,a)$, then the term $\inf_{\cM \in \cM_{\epsilon}} \mathbb{E}_{s^{\prime} \sim T_{\mathcal{M}}\left(s^{\prime} \mid s, a\right)} [ \max_{a'\in \cA} Q(s', a')]$ can be equivalently written as following equation according to Eq. \eqref{aeq_1}.

\begin{equation*}
	\sup_{\lambda \geq 0}\left \{ \mathbb{E}_{s'\sim T_{\cM^\star}} \left[
	\inf_{\bar{s} \in \cS} \left(\max_{a'\in \cA}Q(\bar{s},a')+\lambda d(s',\bar{s})^p  \right)
	-\lambda\epsilon\right]\right\}.
\end{equation*}
This completes the proof.

\subsection{Proof of Theorem \ref{the_1}}\label{proof_the1}

\begin{lemma}
	\label{lem_2}
	For any $\pi \in \Pi$, the performance between the true MDP $\mathcal{M}^\star$ and the estimated MDP $\widehat{\mathcal{M}}$ under policy $\pi$ satisfies:
	\begin{equation}
		J(\cM^\star,\pi)-J(\widehat{\cM},\pi)\leq \frac{R_{max}\mathscr{C}_{\cM_{\epsilon}}(\pi)}{(1-\gamma)^2}\sqrt{\frac{2\log \left(|\cM|/\delta\right)}{n}}.
	\end{equation}
   where $\mathscr{C}_{\cM_{\epsilon}}(\pi)$ is defined in Eq. \eqref{def}, $|\cM|$ denote the cardinal number of $\cM$.
\end{lemma}
\noindent \textbf{\emph{Proof of Lemma \ref{lem_2}.}} The proof follows \cite{yu2020mopo}. Let $\mathcal{U}_i(\cM^\star,\widehat{\cM},\pi)$ denote the expectation of cumulative discounted reward under the estimated MDP $\widehat{\cM}$ for $i$ steps and then switching to the true MDP ${\cM^\star}$ for the rest steps. That is,
\begin{equation}
	\label{mopo}
	\mathcal{U}_i(\cM^\star,\widehat{\cM},\pi)=\underset{\substack{\forall s\sim \mu_0 \\ \forall t \geq 0, a_t \sim \pi\left(\cdot \mid s_t\right) \\ \forall i>t \geq 0, s_{t+1} \sim T_{\widehat{\cM}}\left(\cdot \mid s_t, a_t\right) \\ \forall t \geq i, s_{t+1} \sim T_{\cM^\star}\left(\cdot \mid s_t, a_t\right)}}{\mathbb{E}}\left[\sum_{t=0}^{\infty} \gamma^t r\left(s_t, a_t\right) \mid s_0=s\right].
\end{equation}
Note that $\mathcal{U}_0 = J(\cM^\star,\pi)$ and $\mathcal{U}_\infty = J(\widehat{\cM},\pi)$, so the following equation can be obtained:
\begin{equation}
	J(\cM^\star,\pi)-J(\widehat{\cM},\pi)=\sum_{i=0}^{\infty}\left(\mathcal{U}_{i+1}-\mathcal{U}_i\right).
\end{equation}
The Equation \eqref{mopo} can be written as
\begin{equation}
	\begin{aligned}
		\mathcal{U}_i(\cM^\star,\widehat{\cM},\pi)=R_i+\underset{s_i, a_i \sim \pi, T_{\widehat{\cM}}}{\mathbb{E}}\left[\underset{s_{i+1} \sim T_{\cM^\star}}{\mathbb{E}}\Big[\gamma^{i+1} V_{\cM^\star}^\pi\left(s_{i+1}\right)\Big]\right] \\
		\mathcal{U}_{i+1}(\cM^\star,\widehat{\cM},\pi)=R_i+\underset{s_i, a_i \sim \pi, T_{\widehat{\cM}}}{\mathbb{E}}\left[\underset{s_{i+1} \sim T_{\widehat{\cM}}}{\mathbb{E}}\Big[\gamma^{i+1} V_{\cM^\star}^\pi\left(s_{i+1}\right)\Big]\right].
	\end{aligned}
\end{equation}
where $R_i$ is the expected reward obtained from the first step to $i$ step under $\pi$ and $\widehat{\cM}$. Then
\begin{equation}
	\begin{aligned}
		\label{U}
		\mathcal{U}_{i+1}-\mathcal{U}_i&=\gamma^{i+1}\mathbb{E}_{s_i, a_i \sim \pi, T_{\widehat{\cM}}}\bigg\{\mathbb{E}_{s_{i+1} \sim T_{\cM^\star}}\Big[V_{\cM^\star}^\pi\left(s_{i+1}\right)\Big]-\mathbb{E}_{s_{i+1} \sim T_{\widehat{\cM}}}\Big[V_{\cM^\star}^\pi\left(s_{i+1}\right)\Big]\bigg\}\\
		&\leq \frac{\gamma^{i+1}R_{max}}{1-\gamma}\mathbb{E}_{s_{i},a_i \sim \pi, T_{\widehat{\cM}}}\bigg[\text{TV}\Big(T_{\mathcal{M}^{\star}}(\cdot \mid s_i, a_i), T_{\widehat{\mathcal{M}}}(\cdot \mid s_i, a_i)\Big)\bigg].
	\end{aligned}
\end{equation}
The equality comes from the fact that $V_{\cM}^\pi\leq R_{max}/(1-\gamma)$. In addition, \cite{agarwal2020flambe,kim2023model} give that with probability at least $1-\delta$, the following equality holds,
\begin{equation}
	\label{TV}
\mathbb{E}_{(s,a) \sim \mathcal{D}}\bigg[\text{TV}\Big(T_{\mathcal{M}^{\star}}(\cdot \mid s, a), T_{\widehat{\mathcal{M}}}(\cdot \mid s, a)\Big)\bigg]\leq\sqrt{\frac{2\log \left(|\cM|/\delta\right)}{n}}.
\end{equation}
Combining Eqs. \eqref{def}, \eqref{U} and \eqref{TV}, we can conclude
\begin{equation}
	\begin{aligned}
		J(\cM^\star,\pi)-J(\widehat{\cM},\pi)&\leq \frac{R_{max}}{1-\gamma}\sum_{i=0}^{\infty}\bigg\{\gamma^{i+1}\mathbb{E}_{s_{i},a_i \sim \pi, T_{\widehat{\cM}}}\bigg[\text{TV}\Big(T_{\mathcal{M}^{\star}}(\cdot \mid s_i, a_i), T_{\widehat{\mathcal{M}}}(\cdot \mid s_i, a_i)\Big)\bigg]\bigg\}\\
		&=\frac{R_{max}}{(1-\gamma)^2}\mathbb{E}_{(s,a) \sim d_{\widehat{\cM}}^\pi(s,a)}\bigg[\text{TV}\Big(T_{\mathcal{M}^{\star}}(\cdot \mid s, a), T_{\widehat{\mathcal{M}}}(\cdot \mid s, a)\Big)\bigg]\\
		&\leq \frac{R_{max}\mathscr{C}_{\cM_{\epsilon}}(\pi)}{(1-\gamma)^2}\mathbb{E}_{(s,a) \sim d_{\cM^\star}^\mu(s,a)}\bigg[\text{TV}\Big(T_{\mathcal{M}^{\star}}(\cdot \mid s, a), T_{\widehat{\mathcal{M}}}(\cdot \mid s, a)\Big)\bigg]\\
		&\leq \frac{R_{max}\mathscr{C}_{\cM_{\epsilon}}(\pi)}{(1-\gamma)^2}\sqrt{\frac{2\log \left(|\cM|/\delta\right)}{n}}.
	\end{aligned}
\end{equation}
This completes the proof.

\begin{lemma}[Lemma 2 in \protect\cite{liu2023domain}]
	\label{lem_3}
	For any $\pi \in \Pi$, the performance between the true MDP $\mathcal{M}^\star$ and the empiral MDP $\overline{\mathcal{M}}$ under policy $\pi$ satisfies:
	\begin{equation}
		\bigg|J(\cM^\star,\pi)-J(\overline{\cM},\pi)\bigg|
			\leq \frac{1}{1-\gamma} \mathbb{E}_{(s, a) \sim d_{\overline{\mathcal{M}}}^\pi(s, a)}\left[\frac{C_{r, \delta}+R_{\max } C_{T, \delta}}{\sqrt{|\mathcal{D}(s, a)|}}\right].
	\end{equation}
where  $|\mathcal{D}(s, a)|$  represents the cardinality of a specific state-action pair  $(s, a)$  in the dataset  $\mathcal{D}$, $C_{r, \delta}$ and $C_{T, \delta}$ are the sampling error coefficients \cite{yu2021combo}.
\end{lemma}

\noindent \textbf{\emph{Proof of Theorem \ref{the_1}.}} Combining the Lemma \ref{lem_2} and \ref{lem_1}, then the below equality holds. 
\begin{equation}
	\label{25}
	\begin{aligned}
		&\bigg|J(\cM_f,\pi)-J(\cM^\star,\pi)\bigg|\\
		=&\bigg|\Big[fJ(\widehat{\cM},\pi)+(1-f)J(\overline{{\cM}},\pi)\Big]-\Big[fJ(\cM^\star,\pi)+(1-f)J(\cM^\star,\pi)\Big]\bigg|\\
		=&f\bigg|J(\widehat{\cM},\pi)-J(\cM^\star,\pi)\bigg|+(1-f)\bigg|J(\overline{\cM},\pi)-J(\cM^\star,\pi)\bigg|\\
		\leq&\frac{fR_{max}\mathscr{C}_{\cM_{\epsilon}}(\pi)}{(1-\gamma)^2}\sqrt{\frac{2\log \left(|\cM|/\delta\right)}{n}}+\frac{1-f}{1-\gamma} \mathbb{E}_{(s, a) \sim d_{\overline{\mathcal{M}}}^\pi(s, a)}\left[\frac{C_{r, \delta}+R_{\max } C_{T, \delta}}{\sqrt{|\mathcal{D}(s, a)|}}\right]
	\end{aligned}
\end{equation}

The final learned policy ${\pi}^\dag$ can be obtained by optimizing the below equation, ${\pi}^\dag =\argmax_{\pi \in \Pi}J(f,\pi)$, where $J(\pi,f)$ is denoted as 
\begin{equation}
	\label{26}
J(\pi,f)=J(\cM_f,\pi)-\frac{f\gamma\beta}{1-\gamma}\eta(\pi).
\end{equation}
Therefore, for any $\pi \in \Pi$, $J(\pi,f)\leq J(\pi^\dag,f)$ holds. Then, combining Eqs. \eqref{25} and \eqref{26}, we can conclude

\begin{equation}
	\begin{aligned}
		&J(\cM^\star,\pi)-J(\cM^\star,\pi^\dag)\\
		\leq &J(\cM_f,\pi)-J(\cM_f,\pi^\dag)+\frac{1-f}{1-\gamma}\bigg[\omega(\pi)+\omega(\pi^\dag)\bigg]+\frac{fR_{max}}{(1-\gamma)^2}\sqrt{\frac{2\log \left(|\cM|/\delta\right)}{n}}\bigg[\mathscr{C}_{\cM_{\epsilon}}(\pi)+\mathscr{C}_{\cM_{\epsilon}}(\pi^\dag)\bigg]\\
		\leq &\frac{f\gamma\beta}{1-\gamma}\bigg[\eta(\pi)-\eta(\pi^\dag)\bigg]+\frac{1-f}{1-\gamma}\bigg[\omega(\pi)+\omega(\pi^\dag)\bigg]+\frac{fR_{max}}{(1-\gamma)^2}\sqrt{\frac{2\log \left(|\cM|/\delta\right)}{n}}\bigg[\mathscr{C}_{\cM_{\epsilon}}(\pi)+\mathscr{C}_{\cM_{\epsilon}}(\pi^\dag)\bigg]
	\end{aligned}
\end{equation}
where $\omega(\pi)=\mathbb{E}_{(s, a) \sim d_{\overline{\mathcal{M}}}^\pi(s, a)}\left[({C_{r, \delta}+R_{\max } C_{T, \delta})}/{\sqrt{|\mathcal{D}(s, a)|}}\right]$. This completes the proof.

\section{Experimental Details} \label{detail}
We conduct experiments on Gym tasks (v2 version), which are included in the D4RL \cite{fu2020d4rl} benchmark. The code for MICRO is available at \href{https://github.com/xiaoyinliu0714/MICRO}{github.com/xiaoyinliu0714/MICRO}. The codes of robust experiment for MOBILE and RAMBO origin from \href{https://github.com/yihaosun1124/mobile}{github.com/yihaosun1124/mobile} and \href{https://github.com/yihaosun1124/OfflineRL-Kit}{github.com/yihaosun1124/OfflineRL-Kit}, respectively.

\subsection{Dynamics Model and Policy Training}
\noindent\textbf{Dynamics Model Training}: Similar to \cite{yu2020mopo}, we train an ensemble of 7 such dynamics models and select the best 5 models. Each model consists of a 4-layer feedforward neural network with 200 hidden units. The model training employs maximum likelihood estimation with a learning rate of $1e-4$ and Adam optimizer.

\vspace{2mm}
\noindent\textbf{Policy Optimization}: Policy optimization is based on the SAC framework \cite{haarnoja2018soft}. The critic network $Q_{\omega}$ and the policy network $\pi_{\theta}$ adopt a 2-layer feedforward neural network with 256 hidden units. The entropy regularization coefficient $\alpha$ in Eq. \eqref{sac} is automatically adjusted, with the entropy target set to -dim(A). The detailed hyperparameters of dynamics model training and policy optimization see table \ref{Base}.

\subsection{Hyperparameters}
The hyperparameters for MICRO consist of three parts: rollout horizon $h \in \{1,5,10\}$, tuning coefficient $\beta \in \{0.1,0.2,0.5,1\}$ and model data proportion $f \in \{0.5,0.95\}$. Table \ref{Hyperparameters} shows the value of hyperparameters in nine tasks.

\begin{table}[H]
	\centering
	\caption{Base hyperparameter configuration of MICRO.}
	\label{Base}
	\renewcommand{\arraystretch}{1}
	%\resizebox{8.5cm}{!}{
		\begin{tabular}{p{0.02cm}p{0.02cm}p{0.25cm}ll}
			\toprule[1pt] 
			&\multicolumn{3}{l}{\quad \textbf{Parameter}}  & \textbf{Value} \\
			\hline		
			\multirow{6}{*}{\rotatebox{90}{\textbf{Environment}}}&\multirow{6}{*}{\rotatebox{90}{\textbf{Model}}}&&Model learning rate & 1e-3 \\
			&&&Number of hidden layers & 4 \\
			&&&Number of hidden units per layer& 200\\
			&&&Batch size & 256 \\
			&&&Number of model networks & 7 \\
			&&&Number of elites & 5 \\
			\hline
			
			\multirow{8}{*}{\rotatebox{90}{\textbf{Policy}}}&\multirow{8}{*}{\rotatebox{90}{\textbf{Learning}}}&
			&Learning rate (policy/critic) & 1e-4/3e-4 \\
			&&&Number of hidden layers & 2 \\
			&&&Number of hidden units per layer& 256\\
			&&&Batch size & 256 \\
			
			&&&Number of critic networks $K$ & 2 \\
			&&&Discount factor $\gamma$ & 0.99 \\
			&&&Soft update parameter $\tau$ & 5e-3 \\
			&&& Number of iterations& 3M\\
			&&&Target entropy & -dim(A) \\

			\bottomrule[1pt]
		\end{tabular}
	%}
\end{table}

\begin{table}[H]
	\small
	\renewcommand{\arraystretch}{0.8}
	\caption{Hyperparameters for MICRO, where Medium-R and Medium-E denote Medium-replay and Medium-expert, respectively.}
	\label{Hyperparameters}
	\centering
	{
		\begin{tabular}{c|ccc|ccc|ccc}
			\toprule[1pt]  
			& \multicolumn{3}{|c|}{Halfcheetah} &\multicolumn{3}{c|}{Hopper}  & \multicolumn{3}{c}{Walker2d} \\
			&Medium& Medium-R&Medium-E&Medium& Medium-R&Medium-E&Medium& Medium-R&Medium-E\\
			\midrule
			$h$&$5$&$5$&$5$&$5$&$5$&$10$&$5$&$1$&$1$\\
			$\beta$&$0.5$&$0.5$&$0.5$&$1$&$0.1$&$1$&$0.5$&$0.2$&$1$\\
			$f$&$0.95$&$0.95$&$0.5$&$0.95$&$0.5$&$0.95$&$0.95$&$0.5$&$0.95$\\
			\bottomrule[1pt]
		\end{tabular}
	}
\end{table}

\section{Additional Experimental Results} 
\subsection{Additional Results for D4RL} \label{D4RL Experiment}
The normalized score is commonly used to evaluate the performance of offline algorithms \cite{fu2020d4rl}. Here, we denote the expected return of a random policy, an expert policy, and the trained policy by offline RL algorithms as $S_r$ (reference min score), $S_e$ (reference max score), and $S$, respectively. Then, the normalized score $\tilde{S}$ is computed by
\begin{equation*}
\tilde{S}=\frac{S-S_r}{S_e-S_r} \times 100=\frac{\text{expected return}-\text{reference min score}}{\text{reference max score}-\text{reference min score}} \times 100.
\end{equation*}

\begin{figure}[H]
	\centering
	\includegraphics[width=17cm]{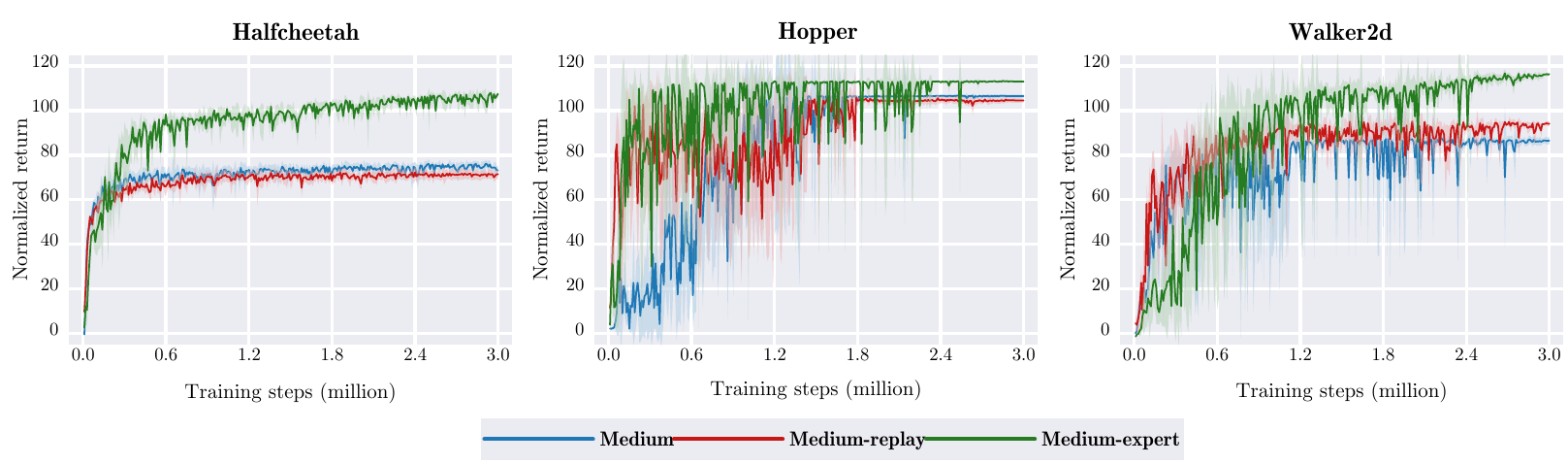}
	\caption{Corresponding learning curves for training. Each figure shows the training curve for a specific task under different datasets.}
	\label{fig4}
\end{figure}

Fig. \ref{fig4} gives the change curves of training in three environments. Each figure consists of three datasets in special environment. It shows the performance of MICRO is gradually improved as the training progresses and eventually converges to a stable value.

\subsection{Additional Results for Robustness} \label{Robust Experiment}
\begin{figure}[H] 
	\centering
	\subfigure[Hopper environment] { \label{fig:a}
		\includegraphics[width=8.6cm]{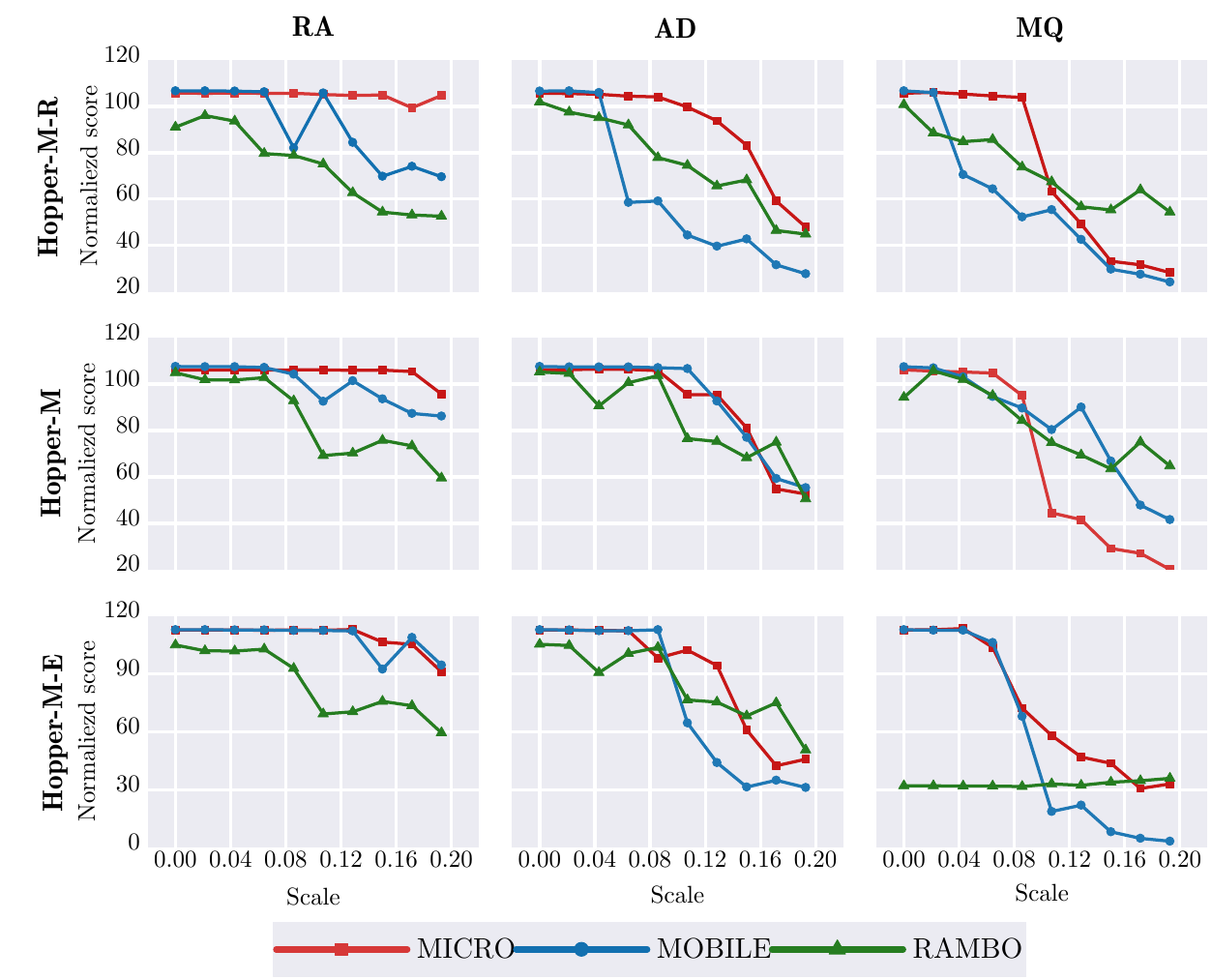}
	}
	\subfigure[Halfcheetah environment] { \label{fig:b}
		\includegraphics[width=8.6cm]{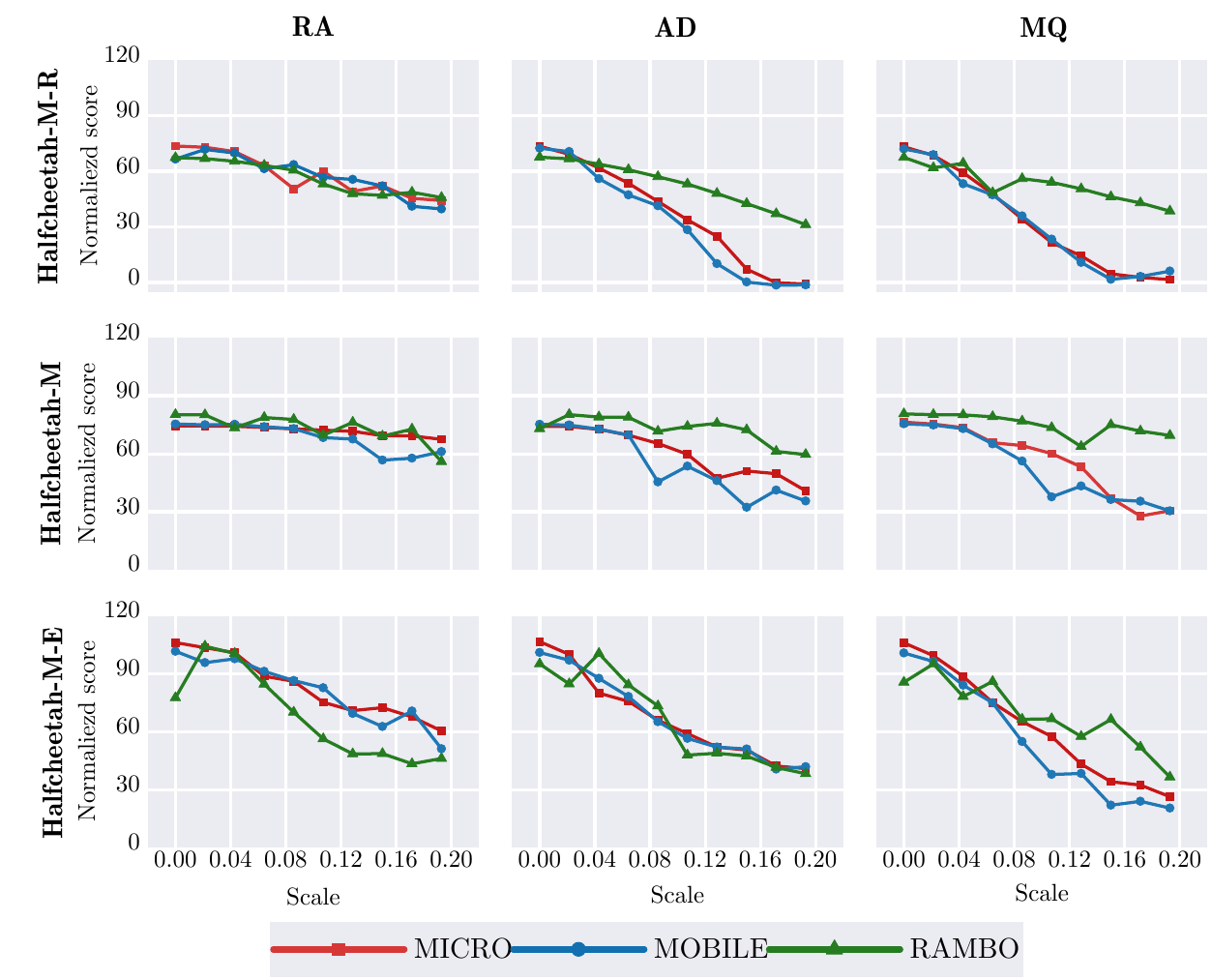}
	}
	\caption{ The performance of MICRO, MOBILE and RAMBO under attack scales range [0, 0.2] of different attack types in three Walker2d datasets. M, M-R and M-E are the abbreviations of Medium, Medium-Replay and Medium-Expert, respectively.}
	\label{fig}
\end{figure}

Fig. \ref{fig} presents the results of different methods with varying perturbation scale $\epsilon$ under three adversarial attacks in Halfcheetah and Hopper environments. From this figure, the performance of the MICRO is superior to MOBILE under three attacks in Hopper and Halfcheetah environments. However, the performance of the MICRO is inferior to RAMBO under AD and MQ attacks in the Halfcheetah environment. RAMBO trains the adversarial model constantly during policy learning, which can also enhance agent robustness. The drawbacks of RAMBO are that it demands more computation cost and the score in the offline benchmark is inferior to the current SOTA methods. The MICRO reduces computation costs significantly and improves the performance in the offline baseline.

\section{Further Discussion}

\textbf{The necessity of less computation and the demand of real-world:} 1) Since the complexities of tuning and the instability of trained policy, less computation can greatly save time cost and computation cost, which is the demand of RL algorithm. 2) In real-world scenarios, such as autonomous driving and medical robots, the demand is to guarantee good performance and high robustness simultaneously. This is the original motivation of MICRO. 3) The large computation is the drawback of previous works, such as RAMBO and ARMOR, which increase time and computation cost. In this paper, less computation doesn't come from real-world demand. It is the demand of algorithm, and time and computation cost.
\vspace{8pt}

\noindent\textbf{The more explanation for parameter $\beta$:}
The ideal $f(s,a)$ is $f(s,a)=\max_{a'\in \mathcal{A}} Q(s', a')-\inf_{\bar{s} \in U_{\epsilon}(s')}[\max_{a'\in \mathcal{A}}Q(\bar{s},a')]$. The term $\mathcal{X}(s,a)$ is used to approximate $U_{\epsilon}(s')$, but it fails to cover all region of the state uncertain set.
The $\beta$ is used to bridge the gap between $U_{\epsilon}(s')$ and $\mathcal{X}(s,a)$. The estimated uncertainty set $\mathcal{X}(s,a)$ is affected by the trained dynamics model. The dynamics model is trained from offline dataset. Concluded from experiments in paper, the possible systematic way for tuning $\beta$ is given below: $\beta$ should be set large in narrow dataset, such as medium and medium-expert, and set small in diverse dataset, such as medium-replay. However, for different datasets, parameter $\beta$ needs to be continuously tuned to guarantee the stability and performance of the agent.
\vspace{8pt}

\noindent\textbf{The connection with uncertainty quantification:} The previous methods, such as MOPO, MoReL, apply reward penalties for OOD data by uncertainty quantification to perform conservative optimization. MICRO penalties $Q$ value through $f(s,a)$ to incorporate conservatism into algorithm, where $f(s,a)$ can be regarded as a type of uncertainty quantification. However, $f(s,a)$ reflects the gap of $Q$ value between the worst-case and average level and contains more robustness information, including the state uncertainty set, the optimal action selected from policy and the worst-case $Q$ value.
\vspace{8pt}

\noindent \textbf{The effectiveness of MICRO:} The main purpose of MICRO is to trade off performance and robustness. Better performance doesn't mean high robustness and high robustness makes performance difficult to improve since robustness considers the worst case. Although it performs poorly than RAMBO under large AD and MQ attacks, MICRO performs greatly better than RAMBO with less computation cost in D4RL tasks and is more robust than MOBILE with comparable performance. The penalty term $f(s,a)$ for model data can be adjusted adaptively according to state uncertainty set. The conservatism of MICRO is mild. However, RAMBO minimizes $Q$ value of model data through model gradient, which is over-conservative and hinders performance improvement. More works should be done to better trade off performance and robustness in the future.
\end{appendices}

	%\twocolumn
	
\end{document}